\documentclass[11pt]{article}
\usepackage{fullpage}
\usepackage{amsmath,bm}
\usepackage{moresize}
\usepackage{amssymb}
\usepackage{tikz-cd}
\usepackage{graphicx}
\usepackage{parskip}
\usepackage{enumitem}
\usepackage{amsthm}
\usepackage{bbm}
\theoremstyle{definition}
\usepackage{listings}
\usepackage{color}
\usepackage{cancel}
\usepackage{framed}
\usepackage{relsize}
\usepackage{yfonts}
\usepackage{anyfontsize}
\usepackage{t1enc}
\usepackage[ruled,vlined]{algorithm2e}
\usepackage{algpseudocode}
\allowdisplaybreaks
\usepackage[utf8]{inputenc}
\usepackage[english]{babel}
\usepackage{lipsum}
\usepackage{mathtools}
\usepackage{verbatim}
\usepackage{multirow}
\usepackage{subfigure}
\usepackage[colorlinks=true]{hyperref}

\usepackage{cleveref}
\crefformat{section}{\S#2#1#3} 
\crefformat{subsection}{\S#2#1#3}
\crefformat{subsubsection}{\S#2#1#3}

\DeclarePairedDelimiter\ceil{\lceil}{\rceil}

\hypersetup{final}

\newtheorem{Thm}{Theorem}

\newtheorem{Lemma}[Thm]{Lemma}
\newtheorem{Def}[Thm]{Definition}

\numberwithin{equation}{section}

\newcommand{\N}{\mathbb{N}}
\newcommand{\Z}{\mathbb{Z}}
\newcommand{\Q}{\mathbb{Q}}
\newcommand{\R}{\mathbb{R}}

\newcommand{\E}{\mathbb{E}}

\newcommand{\eq}{\equiv}

\newcommand{\rpm}{\raisebox{.2ex}{$\scriptstyle\pm$}}

\newcommand{\Cc}{\mathcal{C}}

\newcommand{\Ff}{\mathcal{F}}

\newcommand{\Pp}{\mathcal{P}}

\newcommand{\X}{\mathcal{X}}

\newcommand{\eb}{\bold{e}}

\newcommand{\Ub}{\bold{U}}

\newcommand{\Vb}{\bold{V}}

\newcommand{\xb}{\bold{x}}
\newcommand{\yb}{\bold{y}}

\newcommand{\Ab}{\bold{A}}
\newcommand{\Bb}{\bold{B}}

\newcommand{\Eb}{\bold{E}}

\newcommand{\Pb}{\bold{P}}
\newcommand{\Sb}{\bold{S}}
\newcommand{\Zb}{\bold{Z}}

\newcommand{\Wb}{\bold{W}}
\newcommand{\Xb}{\bold{X}}
\newcommand{\Yb}{\bold{Y}}
\newcommand{\Ib}{\bold{I}}
\newcommand{\Sigb}{\bold{\Sigma}}
\newcommand{\Pib}{\bold{\Pi}}
\newcommand{\Omb}{\bold{\Omega}}

\newcommand{\At}{\tilde{A}}
\newcommand{\Abt}{\tilde{\bold{A}}}
\newcommand{\gamt}{\tilde{\gamma}}
\newcommand{\ellb}{\bar{\ell}}

\newcommand{\blue}{\color{blue}}
\newcommand{\cyan}{\color{cyan}}
\newcommand{\purple}{\color{purple}}
\definecolor{LightBlue}{rgb}{0, 0.7, .7}
\definecolor{Green1}{rgb}{0.0, 0.5, 0.0}
\definecolor{green}{rgb}{0.0, 0.42, 0.24}
\definecolor{byzantine}{rgb}{0.74, 0.2, 0.64}

\algrenewcommand{\Return}{\State\algorithmicreturn~}

\newcommand{\ind}{\text{\color{white}.$\quad$}}

\newcommand{\FFSVD}{\textsf{FastFrobeniusSVD}}
\newcommand{\RanSam}{\textsf{RandomizedSampling}}

\makeatletter
\def\legendre@dash#1#2{\hb@xt@#1{%
  \kern-#2\p@
  \cleaders\hbox{\kern.5\p@
    \vrule\@height.2\p@\@depth.2\p@\@width\p@
    \kern.5\p@}\hfil
  \kern-#2\p@
  }}
\def\@legendre#1#2#3#4#5{\mathopen{}\left(
  \sbox\z@{$\genfrac{}{}{0pt}{#1}{#3#4}{#3#5}$}%
  \dimen@=\wd\z@
  \kern-\p@\vcenter{\box0}\kern-\dimen@\vcenter{\legendre@dash\dimen@{#2}}\kern-\p@
  \right)\mathclose{}}
\newcommand\legendre[2]{\mathchoice
  {\@legendre{0}{1}{}{#1}{#2}}
  {\@legendre{1}{.5}{\vphantom{1}}{#1}{#2}}
  {\@legendre{2}{0}{\vphantom{1}}{#1}{#2}}
  {\@legendre{3}{0}{\vphantom{1}}{#1}{#2}}
}
\def\dlegendre{\@legendre{0}{1}{}}
\def\tlegendre{\@legendre{1}{0.5}{\vphantom{1}}}
\makeatother

\DeclareMathOperator*{\argmin}{arg\,min}

\title{Dimensionality Reduction for $k$-means Clustering}
\author{Neophytos Charalambides}
\date{April 27, 2020}

\begin{document}

\maketitle

\begin{abstract}
We present a study on how to effectively reduce the dimensions of the $k$-means clustering problem, so that provably accurate approximations are obtained. Four algorithms are presented, two \textit{feature selection} and two \textit{feature extraction} based algorithms, all of which are randomized. These algorithms are taken from \cite{BDM09}, \cite{BZD10} and \cite{BZMD14}.
\end{abstract}

\maketitle

\section{Introduction}
\label{Intro}

$\ind$ There is no doubt that \textit{clustering} is a task in high demand throughout many fields, ranging from bioinformatics, to image analysis and data compression. Along with modern developments and the necessity of large high-dimensional datasets, which due to their nature result in \textit{overfitting} of many machine learning algorithms, it is crucial that one reduces the complexity of the algorithms involving these datasets. The authors of \cite{BDM09} are of the first to address the issue of clustering in such datasets with provably accurate approximation results, by proposing a simple pre-processing step to the ``$k$-means'' clustering algorithm; also known as Lloyd's method \cite{Llo82} --- probably the most widely used and popular clustering algorithm. Other references which followed this series of work are \cite{DF09}, \cite{BZD10}, \cite{BZMD14} and \cite{CEMMP15}.\\
$\ind$ Though simple, the $k$-means clustering problem is an NP-hard optimization problem, which means it is very unlikely that there exists an efficient algorithm that solves it \cite{DF09}, \cite{ADHP09}, \cite{MNPV12}. This holds true even for the simplest case when $k=2$ \cite{DFKVV04}. Approximation algorithms to $k$-means have been around for a while; e.g. \cite{KSS04} and \cite{AV06}, though algorithm \ref{1st_feat_sel_alg} \cite{BDM09} was the first provably accurate \textit{feature selection} algorithm for $k$-means clustering. Feature selection for clustering seeks to identify those features that have the \textit{most discriminative power} among the set of all features. On the other hand, \textit{feature extraction} for clustering seeks to create artificial feature which encompass most of the power (hopefully) of all the original features.\\ 
$\ind$ The $k$-means problem is also referred to as \textit{vector quantization}, originating in signal processing for applications related to data compression \cite{DF09}. The goal of finding the optimal set of centroids $\{\mu_i\}_{i\in\N_k}$ for $\N_k\coloneqq\{1,\cdots,k\}$, is equivalent to finding a representative \textit{codebook}.\\
$\ind$ We present four dimensionality reduction algorithms for the $k$-means clustering problem, along with their theoretical guarantees. Two feature selection algorithms are described in \cref{feat_sel_algs_sec}, and two feature extraction algorithms in \cref{feat_extr_algs_sec}. We start off in \cref{backgr_sec} by first formulating the $k$-means problem in terms of linear algebra, and give some of the necessary dimensionality reduction tools, lemmas and subroutine algorithms which will be needed for analysing our four main algorithms. We finally give some concluding remarks and mention some improvements which have taken place in \cref{sum_concl_sec}.

\section{Background and Preliminaries}
\label{backgr_sec}

\subsection{$k$-means and its Linear-Algebraic Formulation}
\label{k_means_subsec}

$\ind$ We first recall the basic $k$-means problem and algorithm. The main idea of the algorithm is that for a given finite set of points $\Pp=\{\xb_1,\cdots,\xb_n\}\subsetneq\R^d$, one can define an objective or cost function with respect to a specified distance; and seek a partition of $\Pp$ into $k$ disjoint non-empty subsets $C_1,\cdots,C_k$, which minimizes this cost. By $\Cc$ we denote the collection of these clusters. The cost function for distance $d(\xb,\yb)=\|\xb-\yb\|_2$, is usually defined as
$$ \Ff\left(\Pp,\{\mu_i^{(t)}\}_{i\in\N_k}\right) = \sum_{i=1}^k\sum_{\xb\in C_i} d\left(\xb,\mu_i^{(t-1)}\right)^2 \qquad \equiv \qquad \Ff(\Pp,\Cc) = \sum_{\xb\in\Pp}\|\xb-\mu(\xb)\|_2^2 $$
at iteration $t$, for $\{\mu_i^{(t-1)}\}_{i\in\N_k}\subsetneq\R^d$, 
the set of \textit{centroids} determined at the previous iteration
$$ \mu_i = \frac{\sum_{\xb_j\in C_i} \xb_j}{s_i} \qquad \text{ for } \qquad s_i=|C_i| $$
and $\mu(\xb)=\{\mu_i$ s.t. $\xb\in C_i\}$ the centroid of the cluster to which $\xb$ belongs. We drop the indicator of the iteration number where it is clear from the context. The $k$-means algorithm is presented in algorithm \ref{alg_k}, and we also formally define the \textit{$k$-means clustering problem}, whose objective is to compute the optimal $k$-partition of $\Pp$
$$ \Cc_{opt} = \argmin_{\Cc}\big\{\Ff(\Pp,\Cc)\big\}. $$
$\ind$ It is also worth noting that the above objective function depends only on the pairwise distances of the points from the corresponding center point, which is tempting to associate it with the Johnson-Lindenstrauss lemma \cite{JL84}. The main reason is that \textit{if all pair-wise distances are preserved, then all clusterings and partitions of the $n$ points --- hence also the optimal partition --- are preserved by the same factor}. By the Johnson-Lindenstrauss lemma it straightforward that one can project the data down to $O(\log(n)/\varepsilon^2)$ dimensions, and guarantee a clustering error which is not more than a factor of $(1+\varepsilon)$ of the optimal clustering error.

\begin{algorithm}[h]
\label{alg_k}
\SetAlgoLined
\KwIn{$\Pp=\{\xb_1,\cdots,\xb_n\}\subsetneq\R^d$ a finite set of points, $k$ the number of clusters}
\KwOut{partition $\Pp=\bigsqcup\limits_{i=1}^kC_i$}
\textbf{Initialize}: $t=0$, centroids $\left\{\mu_i^{(0)}\right\}_{i=1}^k\subsetneq\R^d$ (preferably at random)\\
  \While {some termination criterion}
  {
    \For{i=1 to n}
     { $c_i^{(t)}\gets \argmin_{\ell\in\N_k}\left\{\left\|\xb_i-\mu_\ell^{(t)}\right\|_2^2\right\}$ \Comment{break ties arbitrarily} }
     Define $C_\ell^{(t)}=\{\xb_i\mid c_i^{(t)}=\ell\}$\\
    \For{j=1 to k}
     {
       $\mu_j^{(t)} \gets \frac{1}{\left|C_j^{(t)}\right|}\cdot\sum\limits_{\xb\in C_j^{(t)}}\xb$
     }
     $t\gets t+1$
  }
 \Return partition of $\Pp$ : $\left\{C_\ell^{(t-1)}\right\}_{\ell\in\N_k}$
 \caption{$k$-means}
\end{algorithm}

\begin{Def}
\textit{The \textbf{indicator matrices} $\Xb\in\R^{n\times k}$ have exactly one non-zero entry per row, which denotes membership. That is, for $i\in\N_n$ and $j\in\N_k$: $\xb_i$ of $\Pp$ belongs to the $j^{th}$ cluster if and only if $\Xb_{ij}\neq0$}.
\end{Def}

Furthermore, the nonzero entries of $\Xb$ are $\Xb_{ij}=1/\sqrt{s_j}$; where $s_j=|C_j|=\|\Xb^{(j)}\|_0$. By $\Xb^{(j)}$ we denote the $j^{th}$ column of $\Xb$, and by $\Xb_{(i)}$ the $i^{th}$ row of $\Xb$. Consequently, the columns of $\Xb$ are normalized and pairwise orthogonal; thus $\Xb^T\Xb=\Ib_k$. Additionally
$$ \Ff(\Ab,\Xb)=\|\Ab-\Xb\Xb^T\Ab\|_F^2 = \sum_{i=1}^n\|\Ab_{(i)}-\Xb_{(i)}\Xb^T\Ab\|_2^2 $$
for $\Ab\in\R^{n\times d}$ the matrix representing $\Pp$; i.e. $\Ab=\big[\xb_1 \cdots \xb_n\big]^T$. We interchange between $\Ff(\Pp,\Cc)$ and $\Ff(\Ab,\Xb)$; as the respective arguments of $\Ff$ represent the same objects. The vectors $\Xb_{(i)}\Xb^T\Ab\in\R^{1\times d}$ match the corresponding centroid that $\{\xb_i\}_{i\in\N_n}$ belong to; for all $i\in\N_n$ \cite{ORSS14}, i.e. $\Xb_{(i)}\Xb^T\Ab=\mu(\xb_i)^T$. Here is a simple example for such an indicator matrix : for $k=3$ and $n=6$, assume that ${\cyan C_1}=\{\xb_1,\xb_3,\xb_4\}$, ${\purple C_2}=\{\xb_2\}$ and ${\blue C_3}=\{\xb_5,\xb_6\}$. It follows that
$$ \Xb = \begin{pmatrix} {\cyan1/\sqrt{3}} & 0 & 0 \\ 0 & {\purple1} & 0 \\ {\cyan1/\sqrt{3}} & 0 & 0 \\ {\cyan1/\sqrt{3}} & 0 & 0 \\ 0 & 0 & {\blue1/\sqrt{2}} \\ 0 & 0 & {\blue1/\sqrt{2}} \end{pmatrix} \in\R_{\geq0}^{n\times k} \qquad \Xb\Xb^T = \begin{pmatrix} {\cyan1/3} & 0 & {\cyan1/3} & {\cyan1/3} & 0 & 0 \\ 0 & {\purple1} & 0 & 0 & 0 & 0 \\ {\cyan1/3} & 0 & {\cyan1/3} & {\cyan1/3} & 0 & 0 \\ {\cyan1/3} & 0 & {\cyan1/3} & {\cyan1/3} & 0 & 0 \\ 0 & 0 & 0 & 0 & {\blue1/2} & {\blue1/2} \\ 0 & 0 & 0 & 0 & {\blue1/2} & {\blue1/2} \end{pmatrix} \in\Q_{\geq0}^{n\times n} $$
and we make the following observations:
\begin{itemize}
  \item $\Xb^T\Xb=\Ib_k$
  \item $(\Xb\Xb^T)_{ij}=1/s_{\ell}$ if $\xb_i,\xb_j\in C_\ell$ and $(\Xb\Xb^T)_{ij}=0$ if $\xb_i,\xb_j$ lie in different clusters
  \item $\Xb\Xb^T$ is symmetric with repeated rows and columns (as indicated with the colors corresponding to the same cluster)
  \item $\|(\Xb\Xb^T)_{(i)}\|_0=s_\ell$, for $\ell$ s.t. $\xb_i\in C_\ell$
  \item $\|(\Xb\Xb^T)_{(i)}\|_1=\|(\Xb\Xb^T)^{(i)}\|_1=1$ for all $i\in\N_n$
  \item $\sum\limits_{i=1}^{n}\sum\limits_{j=1}^{n}(\Xb\Xb^T)_{ij}=\sum\limits_{i=1}^{n}1=n$
  \item $\sum\limits_{i=1}^{n}(\Xb\Xb^T)_{ii}=\sum\limits_{\ell=1}^ks_{\ell}\cdot\frac{1}{s_{\ell}}=k \quad \implies \quad \text{tr}(\Xb\Xb^T)=k$
  \item if $\Pp$ is permuted in such a way that the points in the same cluster are with consecutive indices (or permute $\Xb$ after the clustering); $\Xb\Xb^T$ is block-diagonal, and may be easier to work with. 
\end{itemize}

\begin{Def}[$k$-means clustering problem]
\textit{Given the matrix $\Ab=\big[\xb_1 \cdots \xb_n\big]^T\in\R^{n\times d}$ for $\Pp=\{\xb_i\}_{i\in\N_n}$; and $k\in\Z_+$ the number of clusters, find the indicator matrix $\Xb_{opt}\in\R^{n\times k}$ such that}
$$ \Xb_{opt} = \arg\min_{\Xb\in\X} \left\{\|\Ab-\Xb\Xb^T\Ab\|_F^2\right\} = \arg\min_{\Xb\in\X} \left\{\|(\Ib_n-\Xb\Xb^T)\Ab\|_F^2\right\} $$
\textit{where $\X$ denotes the set of all $n\times k$ indicator matrices $\Xb$. The \textbf{optimal value} of the $k$-means clustering objective is}
\begin{equation}
\label{k_means_obj}
  F_{opt} = \min_{\Xb\in\X}\left\{\|\Ab-\Xb\Xb^T\Ab\|_F^2\right\} = \|\Ab-\Xb_{opt}\Xb_{opt}^T\Ab\|_F^2 \qquad \eq \qquad F_{opt}=\Ff(\Ab,\Xb_{opt}).
\end{equation}
\end{Def}

\begin{Def}[$k$-means $\gamma$-approximation algorithm]
\label{gam_appr_def}
\textit{An algorithm is a \textbf{$\gamma$-approximation} for the $k$-means clustering problem (with $\gamma\geq1$), if it takes as inputs $\Ab$ and $k$, and returns an indicator matrix $\Xb_\gamma$ for which}
$$ \Pr\left[\|\Ab-\Xb_\gamma \Xb_\gamma^T\Ab\|_F^2 \leq \gamma\min_{\Xb\in\X}\left\{\|\Ab-\Xb\Xb^T\Ab\|_F^2\right\}\right] \geq 1-\delta_\gamma $$
\textit{where $\delta_\gamma\in[0,1)$ is the failure probability of the algorithm}.
\end{Def}

\subsection{Dimensionality Reduction and Further Tools}
\label{dim_red_subsec}

$\ind$ The main idea of speeding many machine learning algorithms, is to reduce the complexity of certain linear algebraic operations. A common way of doing so, is to reduce the dimension of the space in which the operations take place, where the cost of doing so is to settle for an approximate solution. In practice this may yield better results, as due to large datasets, a lot of the information may be redundant and we may run into issues such as overfitting.\\
$\ind$ Dimensionality reduction is achieved by projecting the dataset into a space of lower dimension, i.e. apply a (linear) transformation $\Pb:\R^d\to\R^r$ for $r<d$ by using a (maybe randomized) projection $\Pb\in\R^{r\times d}$. By applying $\Pb$ on $\Pp$ we get $\hat{\Pp}=\{\hat{\xb}_1,\cdots,\hat{\xb}_n\}\subsetneq\R^r$ for $\hat{\xb}_i=\Pb\xb_i$. For our purposes, once the new dataset $\hat{\Pp}$ is constructed, we perform $k$-means on it to \textit{approximate} the optimal partition
$$ \hat{\Cc}_{opt} = \argmin_{\Cc}\left\{\Ff(\hat{\Pp},\Cc)\right\} \qquad \text{such that} \qquad \Ff(\Pp,\hat{\Cc}_{opt}) \leq \gamma\cdot\Ff(\Pp,\Cc_{opt}) $$
for some $\gamma>0$. That is, by reducing the dimension and then performing $k$-means; we get a $\gamma$-approximation of the optimal clustering.\\

The more common dimensionality reduction approaches for $k$-means clustering, are:
\begin{itemize}
  \item \textbf{feature selection}: a small subset $r$ of the actual $d$ features from the data are selected.
  \item \textbf{feature extraction}: a small set of $r$ artificial features is constructed for each $d$-dimensional data sample.
\end{itemize}
In \cref{feat_sel_algs_sec} we present two provably accurate feature selection algorithms for $k$-means clustering, and in \cref{feat_extr_algs_sec} two provably accurate feature extractions algorithms for $k$-means clustering.\\

$\ind$ Before we move on to the actual algorithms, we give some notation and lemmata. Most of them are taken from \cite{BDM09} and \cite{BZMD14}, in which the proofs may be found.

\begin{Lemma}[Matrix Pythagorean Theorem]
\label{mat_pyth_thm}
\textit{For $\Wb,\Yb\in\R^{m\times n}$ satisfying $\Wb\Yb^T=\bold{0}_{m\times m}$, we have $\|\Wb+\Yb\|_F^2=\|\Wb\|_F^2+\|\Yb\|_F^2$}.
\end{Lemma}

\textbf{Singular Value Decomposition}: The  singular value decomposition (SVD) of a matrix $\Ab\in\R^{m\times n}$ is a powerful tool used in numerical analysis. Some even describe it as both the ``Swiss Army Knife'' and the ``Rolls-Royce'' of matrix decompositions \cite{GMDL06}, \cite{LKT18}. Assuming the usual conventions of this decomposition, the SVD of $\Ab\in\R^{m\times n}$ with rank$(\Ab)=\rho$ is $\Ab=\Ub_{\Ab}\Sigb_{\Ab}\Vb_{\Ab}^T$, with
$$ \Ab = \overbrace{\begin{pmatrix} \Ub_k & \Ub_{\rho-k} \end{pmatrix}}^{\Ub_{\Ab}\in\R^{m\times \rho}} \overbrace{\begin{pmatrix} \Sigb_k & \bold{0} \\ \bold{0} & \Sigb_{\rho-k} \end{pmatrix}}^{\Sigb_{\Ab}\in\R^{\rho\times \rho}} \overbrace{\begin{pmatrix} \Vb_k^T \\ \Vb_{\rho-k}^T \end{pmatrix}}^{\Vb_{\Ab}^T\in\R^{\rho\times n}} $$
where the sizes of the indicated submatrices are $\Ub_k\in\R^{m\times k}$, $\Ub_{\rho-k}\in\R^{m\times(\rho-k)}$, $\Vb_k\in\R^{n\times k}$, $\Vb_{\rho-k}\in\R^{n\times(\rho-k)}$, $\Sigb_k\in\R^{k\times k}$ and $\Sigb_{\rho-k}\in\R^{(\rho-k)\times(\rho-k)}$. The ordered singular values of $\Ab$ are denoted by $\sigma_i(\Ab)$. Where clear from the context, we drop the subscript indicating the matrix whose SVD parts we are dealing with. A known fact which has many applications throughout numerical linear algebra and machine learning; e.g. PCA, is that
\begin{equation}
\label{opt_A_k}
  \argmin_{\substack{\Bb\in\R^{m\times n}\\ \text{rank}(\Bb)\leq k\leq \rho}} \big\{\|\Ab-\Bb\|_F^2\big\} = \Ab_k = \Ub_k\Sigb_k\Vb_k^T = \Ab\Vb_k\Vb_k^T = \Ub_k\Ub_k^T\Ab.
\end{equation}
The ``$k$'' here is \textit{not} referring to the number of clusters we require by the $k$-means algorithm. We overload the use of ``$k$'' to refer also to the rank approximation of lower-rank approximations of a matrix, as this is how it is referred to in most of the literature. An important decomposition which was used in \cite{DFKV99} to give an upper bound to the approximation of the problem they consider, is
\begin{equation}
\label{sep_id_A}  
  \Ab = \overbrace{\Ab\Vb_k\Vb_k^T}^{\Ab_k}+\overbrace{\Ab-\Ab\Vb_k\Vb_k^T}^{\Ab_{\rho-k}}
\end{equation}
which for the approximate cluster indicator matrix $\hat{\Xb}_{opt}$, results in
$$ \Big(\left(\Ib_m-\hat{\Xb}_{opt}\hat{\Xb}_{opt}^T\right)\Ab_k\Big)\cdot\Big(\left(\Ib_m-\hat{\Xb}_{opt}\hat{\Xb}_{opt}^T\right)\Ab_{\rho-k}\Big)^T = \bold{0}_{m\times m} \qquad \text{and} \qquad \Ab\cdot\Ab_{\rho-k}^T=\bold{0}_{m\times m} $$
for $P_{\Xb}\coloneqq\hat{\Xb}_{opt}\hat{\Xb}_{opt}^T$ and $P_{\Xb^{\perp}}\coloneqq(\Ib_m-\hat{\Xb}_{opt}\hat{\Xb}_{opt}^T)$ projection matrices. Furthermore, $\Ab_k$ is an orthogonal projection of $\Ab$ and $\Ab_{\rho-k}$ is its residual, and by the Pythagorean theorem we have $\|\Ab_k\|_F^2+\|\Ab_{\rho-k}\|_F^2=\|\Ab_k+\Ab_{\rho-k}\|_F^2=\|\Ab\|_F^2$.\\
$\ind$ Constructing the exact SVD takes cubic time, which is not ideal. Below we give a statement for the \textit{existence} of fast relative-error Frobenius norm SVD approximations, and then present such an algorithm. Similar algorithms along with further details may be found in \cite{BDM14} and \cite{Sar06}.
\vspace{5mm}

\begin{Lemma}[\cite{BZMD14}]
\label{FFSVD_lem}
\textit{Given $\Ab\in\R^{n\times d}$ with} rank$(\Ab)=\rho$, \textit{a target rank $2\leq k<\rho$, and error parameter $\varepsilon\in(0,1)$, there exists a randomized algorithm that computes $\Zb\in\R^{d\times k}$ such that}
$$ \Zb^T\Zb=\Ib_k \ \ \text{ and } \ \ \Eb\Zb=\bold{0}_{n\times k} \quad \text{for} \quad \Eb=\Ab-\Ab\Zb\Zb^T=\Ab(\Ib_d-\Zb\Zb^T)\in\R^{n\times d} $$
\begin{equation}
\label{FFSVD_lem_eq}
  \text{and} \qquad \E\big[\|\Eb\|_F^2\big]\leq(1+\varepsilon)\cdot\|\Ab-\Ab_k\|_F^2 = (1+\varepsilon)\cdot\sqrt{\sum_{i=k+1}^\rho\sigma_i^2(\Ab)} \ .
\end{equation}
\textit{We denote the proposed algorithm by} $\Zb=\FFSVD(\Ab,k,\varepsilon)$ \textit{which takes time} $O(ndk/\varepsilon)$.
\end{Lemma}

$\ind$ We point out that in lemma \ref{FFSVD_lem_eq} we retrieve the rank-$k$ matrix $\Ab\Zb\Zb^T$, which is almost as good an approximation to $\Ab$ (in expectation) with $\Ab_k$. By comparing it with $\Ab_k=\Ab\Vb_k\Vb_k^T$, we conclude that $\Zb\simeq\Vb_k$.

\begin{algorithm}[H]
\label{FFSVD_alg}
\KwIn{$\Ab\in\R^{n\times d}$, $k\in[2,\rho)$ the rank of the approximation, $\varepsilon\in(0,1)$ an error parameter}
\KwOut{$\Zb\in\R^{d\times k}$ and approximation to $\Vb_k$}
\begin{enumerate}
  \item $r \gets k+\ceil{k/\varepsilon+1}$
  \item Generate $\bold{R}\sim \mathcal{N}(0,1)$ a standard normal Gaussian matrix of size $d\times r$, with i.i.d. entries
  \item $\Yb\gets \Ab\bold{R}\in\R^{n\times r}$
  \item Orthonormalize the columns of $\Yb$ to construct $\bold{Q}\in\R^{n\times r}$
  \item Let $\Zb\in\R^{d\times k}$ be the top $k$ right singular vectors of $\bold{Q}^T\Ab\in\R^{r\times d}$
\end{enumerate}
 \Return $\Zb$
 \caption{$\FFSVD(\Ab,k,\varepsilon)$}
\end{algorithm}

$\ind$ We also give algorithm \ref{rand_sampl_alg}, a randomized sampling algorithm which will be used as a subroutine in one of the dimensionality reduction $k$-means algorithms we will present. The main idea is that the rows of the input matrix comprise the data samples of dimension $d$; from which we sample based on a distribution defined in the algorithm, and then rescale the sampled rows accordingly. We refer to this algorithm as $\RanSam$, which takes $O(nd)$ time to compute the distribution $\{p_i\}_{i\in\N_n}$, and $O(n+r)$ time to implement the sampling procedure via the technique in \cite{Vos91}. In total, $\RanSam(\Ab,r)$ takes $O(nd)$ time.

\begin{algorithm}[H]
\label{rand_sampl_alg}
\KwIn{$\Ab\in\R^{n\times d}$ with $n>k$, and integer parameter $r<n$ the number of sampled rows}
\KwOut{Sampling matrix $\Omb\in\{0,1\}^{n\times r}$, and diagonal rescaling matrix $\Sb\in\R^{r\times r}$}
\textbf{Initialize}: $\Omb=\bold{0}_{n\times r}$ and $\Sb=\bold{0}_{r\times r}$
\begin{enumerate}
  \item For all $i\in\N_n$ define $p_i=\|\Ab_{(i)}\|_2^2/\|\Ab\|_F^2$ \Comment{$\sum_{i=1}^np_i=1$}
  \item \For{$t=1$ to $r$}
      {
        $\quad \ $ Sample $i_t$ from $\N_n$ based on the distribution $\{p_i\}_{i\in\N_n}$\\
        $\quad \ $ $\Omb_{i_t,t}=1$ \Comment{equivalently $\Omb^{(t)}=\eb_{i_t}$}\\
        $\quad \ $ $\Sb_{t,t=1/\sqrt{rp_{i_t}}}$
      }
\end{enumerate}
 \Return $[\Omb,\Sb]$
 \caption{$\RanSam(\Ab,r)$}
\end{algorithm}

$\ind$ Lastly, we present six lemmas before we proceed to the main algorithms. The lemmas are self contained and explanatory, though if one wishes further details on the intuition or the significance and importance of the lemmas, as well as their proofs, she or he may refer to \cite{BDM09}, \cite{BZD10} and \cite{BZMD14}.
\vspace{5mm}

\begin{Lemma}[\cite{BZMD14}]
\label{lem_1}
\textit{Let $\Vb\in\R^{n\times k}$ with $n>k$ and $\Vb^T\Vb=\Ib_k$, $\delta\in(0,1)$, $r\in(4k\ln(2k/\delta),n]$, and} $[\Omb,\Sb]=\RanSam(\Vb,r)$. \textit{Then, for all $i=1,\cdots,k$}
$$ \Pr\left[1-\sqrt{\frac{4k\ln(2k/\delta)}{r}}\leq\sigma_i^2(\Vb^T\Omb\Sb)\leq1+\sqrt{\frac{4k\ln(2k/\delta)}{r}}\right]\geq 1-\delta. $$
\end{Lemma}
\vspace{5mm}

\begin{Lemma}[\cite{BZMD14}]
\label{lem_2}
\textit{For any $r\geq1$, $\Wb\in\R^{n\times k}$ and $\Yb\in\R^{m\times n}$, let} $[\Omb,\Sb]=\RanSam(\Wb,r)$. Let $\delta\in(0,1)$. \textit{Then
$$ \Pr\left[\|\Yb\Omb\Sb\|_F^2\leq\frac{1}{\delta}\|\Yb\|_F^2\right]\geq 1-\delta. $$}
\end{Lemma}
\vspace{5mm}

\begin{Lemma}[\cite{BZMD14}]
\label{lem_3}
\textit{Fix $\Ab\in\R^{n\times d}$, $k\geq1$, $\varepsilon\in(0,1/3)$, $\delta\in(0,1)$ and $r=4k\ln(2k/\delta)/\varepsilon^2$. Compute the matrix $\Zb\in\R^{d\times k}$ from lemma \ref{FFSVD_lem}; such that $\Ab=\Ab\Zb\Zb+\Eb$, i.e.} $\Zb=\FFSVD(\Ab,k,\varepsilon)$. \textit{Then run} $[\Omb,\Sb]=\RanSam(\Zb,r)$, \textit{and we have}
$$ \Pr\left[\exists\tilde{\Eb}\in\R^{n\times d} \text{ s.t. } \Ab\Zb\Zb^T=\big(\Ab\Omb\Sb(\Zb^T\Omb\Sb)^{\dagger}\Zb^T+\tilde{\Eb}\big) \text{ and } \|\tilde{\Eb}\|_F\leq\frac{1.6\varepsilon}{\sqrt{\delta}}\|\Eb\|_F \right] \geq 1-3\delta. $$
\end{Lemma}
\vspace{5mm}

\begin{Lemma}[\cite{BDM09}]
\label{lem_4}
\textit{Assume that for $\Ab\in\R^{n\times d}$, parameters $\varepsilon\in(0,1)$ and $k$ the number of clusters, we retrieve $\Sb$ and $\Omb$ from algorithm \ref{1st_feat_sel_alg}. Let $c_0$ and $c_1$ be absolute constants, based on} \cite{RV07} \textit{theorem 3.1 (take them to be ``sufficiently large'')\footnote{The experimental results from \cite{BZMD14} indicate that the large constants which appear in algorithms \ref{2nd_feat_sel_alg} and \ref{1st_feat_extr_alg} are artifacts of the theoretical analysis, and can be (certainly) improved.}. If the sampling parameter $r$ satisfies
$$ r\geq\frac{2c_1c_0^2k}{\varepsilon^2}\cdot\log(\frac{c_1c_0^2k}{\varepsilon^2})\eqqcolon\zeta $$
then the following four statements all hold simultaneously, with probability at least $1/2$}:
\begin{enumerate}
  \item $\|\Vb_k^T\Omb\Sb\|_2=\sigma_{max}(\Vb_k^T\Omb\Sb)\leq\sqrt{1+\alpha}$
  \item $\|(\Vb_k^T\Omb\Sb)^{\dagger}\|_2=1/\sigma_{min}(\Vb_k^T\Omb\Sb)\leq\sqrt{1/(1-\alpha)}$
  \item rank$(\Vb_k^T\Omb\Sb)=k$, \textit{i.e. it is full-rank}
  \item $\Ab_k=(\Ab\Omb\Sb)(\Vb_k^T\Omb\Sb)^{\dagger}\Vb_k^T+\Eb$, with $\|\Eb\|_F\leq \beta\|\Ab-\Ab_k\|_F$
\end{enumerate}
\textit{where to simplify notation, we set $\alpha=6\varepsilon/\sqrt{c_1}$ and $\beta=\sqrt{6/\zeta}+\sqrt{6\alpha^2/(1-\alpha)}$}.
\end{Lemma}
\vspace{5mm}

\begin{Lemma}[\cite{BZMD14}]
\label{lem_5}
\textit{Fix any matrix $\Yb\in\R^{n\times d}$, $k>1$ and $\varepsilon>0$. Let $\Pib\in\R^{d\times r}$ be a rescaled random sign matrix with $r=c_0k/\varepsilon^2$; for $c_0\geq100$, with entries
$$ \Pib_{ij} = \begin{cases} +1/\sqrt{r} \qquad \text{w.p. } 1/2 \\ -1/\sqrt{r} \qquad \text{w.p. } 1/2 \end{cases}. $$
Then
$$ \Pr\Big[\|\Yb\Pib\|_F^2\geq(1+\varepsilon)\|\Yb\|_F^2\Big]\leq1/100. $$
}
\end{Lemma}
\vspace{5mm}

\begin{Lemma}[\cite{BZMD14}]
\label{lem_6}
\textit{Let $\Ab\in\R^{n\times d}$ with} $\rho=$rank$(\Ab)<k$, $\Ab_k=\Ub_k\Sigb_k\Vb_k^T$ \textit{and $\varepsilon\in(0,1/3)$. Let $\Pib\in\R^{d\times r}$ with $r=c_0k/\varepsilon^2$; for $c_0\geq3330$, be a (rescaled) random sign matrix with entries
$$ \Pib_{ij} = \begin{cases} +1/\sqrt{r} \qquad \text{w.p. } 1/2 \\ -1/\sqrt{r} \qquad \text{w.p. } 1/2 \end{cases}. $$
The following then simultaneously hold, with probability at least $0.97$:
\begin{enumerate}
  \item For all $i\in\N_k$: $(1-\varepsilon)\leq\sigma_i^2(\Vb_k^T\Pib)\leq(1+\varepsilon)$
  \item There exists a $\tilde{\Eb}\in\R^{n\times d}$ such that:
  $$ \Ab_k=\Ab\Pib(\Vb_k^T\Pib)^{\dagger}\Vb_k^T+\tilde{\Eb} \quad \text{ and } \quad \|\tilde{\Eb}\|_F\leq3\varepsilon\|\Ab-\Ab_k\|_F. $$
\end{enumerate}}
\end{Lemma}

\section{Randomized Feature Selection Algorithms}
\label{feat_sel_algs_sec}

\subsection{First Randomized Feature Selection Algorithm --- Leverage Score Sampling}
\label{1st_sel_alg_sec}

$\ind$ There are many sampling algorithms used in various applications, which similar in spirit to algorithm \ref{rand_sampl_alg}; sample (without replacement) based on the \textit{leverage scores}. The right \textbf{leverage scores} of a matrix are the squared Euclidean norms of the rows of the \textit{reduced} right singular vector matrix $\Vb_\Ab\in\R^{d\times\rho}$ for rank$(\Ab)=\rho$. It is worth noting that the leverage scores are not defined based on the SVD, but rather the matrix $\Ab$, i.e. any orthonormal basis would suffice (e.g. could use a $\bold{Q}\bold{R}$ decomposition instead), and the left leverage score values are in fact the diagonal entries of the projection matrix $P_\Ab=\Ab\Ab^{\dagger}$, i.e. $\ell_i=(P_\Ab)_{ii}$. Furthermore, the right leverage scores of $\Ab$ are the same as the left leverage scores of $\Ab^T$. The left leverage scores characterize the importance of the corresponding data point form the data matrix $\Ab$.\\ 
$\ind$ Since we are interested in feature selection, we instead deal with the \textit{normalized} right leverage scores of $\Ab$
\begin{equation}
\label{norm_lvg_sc}
  \ellb_i \coloneqq \frac{\left\|(\Vb_k)_{(i)}\right\|_2^2}{\|\Vb_k\|_F^2} =  \frac{\left\|(\Vb_k)_{(i)}\right\|_2^2}{k} \qquad \text{ for all } i=1,2,\cdots,k
\end{equation}
which characterize the importance of the corresponding \textit{feature}; with respect to the $k$-means objective \eqref{k_means_obj}. The scores $\left\{\ellb_i\right\}_{i=1}^k$ form a probability distribution, as $\sum_{i=1}^k\ellb_i=1$.\\
$\ind$ The following algorithm is the first provably accurate feature selection algorithm for $k$-means clustering\footnote{There was a discrepancy here, as in \cite{BZMD14} the authors also claim to have presented the first provably accurate feature selection algorithm for $k$-means clustering. Ironically, three authors are mutual in the two papers.} \cite{BDM09}, which dates back to 2009. Its quality-of-approximation result is given in theorem \ref{qual_appr_thm_1}.

\begin{algorithm}[h]
\label{1st_feat_sel_alg}
\SetAlgoLined
\KwIn{$\Ab\in\R^{n\times d}$ for $n$ points and $d$ features, number of clusters $k$, parameter $\varepsilon\in(0,1)$}
\KwOut{$\Abt\in \R^{n\times r}$, with $r=\Theta\left(k\log(k/\varepsilon)/\varepsilon^2\right)$}
Compute $\Vb_k\in\R^{d\times k}$, the top $k$ right singular vectors of $\Ab$\\
Compute the normalized leverage scores $\left\{\ellb_i\right\}_{i=1}^d$ as in \eqref{norm_lvg_sc}\\
  \For{t=1 to r}
    {keep the $i^{th}$ feature with probability $\ellb_i$, and scale it by $1/\sqrt{r\ellb_i}$ \Comment{run i.i.d. random trials}}
 \Return $\Abt\in\R^{n\times r}$, consisting of the selected (rescaled) $r$ feature vectors as its columns
 \caption{Randomized Feature Selection, Based on Leverage Scores \cite{BDM09}}
\end{algorithm}

\begin{Thm}
\label{qual_appr_thm_1}
\textit{Run algorithm \ref{1st_feat_sel_alg} with appropriate $\Ab$, $\varepsilon$ and $r$, to get $\Abt$ in time $O(nd\cdot\min\{n,d\})$. If we run any $\gamma$-approximation algorithm $(\gamma\geq1)$ for the $k$-means clustering problem on inputs $\Abt$ and $k$, whose failure probability is $\delta_\gamma$, the resulting cluster indicator matrix $\Xb_{\gamt}$ satisfies}\footnote{In \cite{BDM09} the theorem claims that the approximation factor is $\big(1+(1+\varepsilon)\gamma\big)$, though \cite{BZMD14} claims that the corresponding error had a fixable bug, and that the factor is in fact $\big(1+(2+\varepsilon)\gamma\big)$. The corrected bound is attained if we change the bound on $\eta_1^2$ from $\gamma(1+\varepsilon)F_{opt}$ to $\gamma(2+\varepsilon)F_{opt}$.}
$$ \Pr\left[\|\Ab-\Xb_{\gamt} \Xb_{\gamt}^T\Ab\|_F^2\leq\big(1+(2+\varepsilon)\gamma\big)\min_{\Xb\in\X}\left\{\|\Ab-\Xb\Xb^T\Ab\|_F^2\right\}\right] \geq \frac{1}{2}-\delta_\gamma. $$
\end{Thm}

$\ind$ Like $\RanSam$, algorithm \ref{1st_feat_sel_alg} can also be defined as the product of a sampling and a rescaling matrix. We perform $r$ trials (without replacement) according to the distribution $\left\{\ellb_i\right\}_{i=1}^k$. The sampling matrix $\Omb\in\{0,1\}^{d\times r}$ has as its $i^{th}$ column the standard basis vector $\eb_{i(j)}$ of length $r$; for $i(j)\in\N_d$ the feature drawn at the $j^{th}$ trial. The rescaling matrix $\Sb\in\R^{r\times r}$ is a diagonal matrix with $\Sb_{ii}=1/\sqrt{r\ellb_{i(j)}}$. The projected matrix is then defined as $\Abt=\Ab\cdot(\Omb\Sb)\in\R^{n\times r}$, i.e. we project $\Ab$ onto the row-span of $\Pib\coloneqq\Omb\Sb\in\R^{d\times r}$.\\
$\ind$ The main idea behind the proof of theorem \ref{qual_appr_thm_1} is to split $\Ab$ into $\Ab_k+\Ab_{\rho-k}$; as in \eqref{sep_id_A}, by the Pythagorean theorem \ref{mat_pyth_thm} we then get
$$ \|\Ab-\Xb_{\gamt}\Xb_{\gamt}^T\Ab\|_F^2 = \overbrace{\|(\Ib_n-\Xb_{\gamt}\Xb_{\gamt}^T)\Ab_k\|_F^2}^{\eta_1^2}+\overbrace{\|(\Ib_n-\Xb_{\gamt}\Xb_{\gamt}^T)\Ab_{\rho-k}\|_F^2}^{\eta_2^2} $$
since the subspaces image$\big(\Ab_k-\Xb_{\gamt}\Xb_{\gamt}^T\Ab_k\big)$ and image$\big(\Ab_{\rho-k}-\Xb_{\gamt}\Xb_{\gamt}^T\Ab_{\rho-k}\big)$ are perpendicular, and then bound the terms $\eta_1$ and $\eta_2$. This resembles the ideas from \cite{DFKVV04}, one of the first provably accurate randomized algorithms.\\
$\ind$ To bound $\eta_1$ various facts are used: lemma \ref{lem_4}, the triangle inequality, the fact that $(\Ib_n-\Xb_{\gamt}\Xb_{\gamt}^T)$ is a projection matrix, the \textit{strong submultiplicativity property} relating $\|\cdot\|_F$ with $\|\cdot\|_2$ (i.e. $\|\Wb\Yb\|_F\leq\|\Wb\|_2\|\Yb\|_F$ for any pair of matrices $\Wb$ and $\Yb$), invariance of norms, and definition \ref{gam_appr_def} is used to replace $\Xb_{opt}$ with $\Xb_{\gamt}$; while also introducing a $\sqrt{\gamma}$ multiplicative factor. The strong submultiplicativity property is also know as \textit{spectral multiplicity}, and it holds because multiplying by a matrix can scale each row or column, and hence the Frobenius norm, by at most the matrix’s spectral norm. This analysis results in the bound \underline{$\eta_1^2\leq\gamma(2+\varepsilon)F_{opt}$}. Thereafter, to bound $\eta_2^2$ we note that
$$ \|\Ab_{\rho-k}\|_F^2=\|\Ab-\Ab_k\|_F^2 \overset{\sharp}{\leq} \|\Ab-\Xb_{opt}\Ab_{opt}^T\Ab\|_F^2 = F_{opt} $$
where $\sharp$ follows from \eqref{opt_A_k}, and since $P_{\Xb_{\gamt}^{\perp}}\coloneqq(\Ib_n-\hat{\Xb}_{\gamt}\hat{\Xb}_{\gamt}^T)$ is a projection matrix
$$ \eta_2^2 = \|P_{\Xb_{\gamt}^{\perp}}\Ab_{\rho-k}\|_F^2 \leq \|P_{\Xb_{\gamt}^{\perp}}\|_F^2\cdot \|\Ab_{\rho-k}\|_F^2 = \|\Ab_{\rho-k}\|_F^2 $$
thus \underline{$\eta_2^2\leq F_{opt}$}. All in all, we get \underline{$\eta_1^2+\eta_2^2\leq \big(1+(1+\varepsilon)\gamma\big)F_{opt}$}. Theorem \ref{qual_appr_thm_1} fails only if lemma \ref{lem_4} or the $\gamma$-approximation $k$-means clustering algorithms fail, which happens with probability at most $1/2+\delta_\gamma$. The time it takes to run algorithm \ref{1st_feat_sel_alg} is $O(nd\cdot\min\{n,d\})$, which is the time it takes to construct $\Vb_k$; in order to compute the leverage scores.

\subsection{Second Randomized Feature Selection Algorithm --- Randomized Sampling}
\label{2nd_sel_alg_sec}

$\ind$ The next randomized feature selection algorithm we present for $k$-means clustering, uses the $\FFSVD(\cdot,\cdot)$ and $\RanSam(\cdot,\cdot)$ algorithms as subroutines. Furthermore, one could replace the first step of algorithm \ref{2nd_feat_sel_alg} with the exact SVD of $\Ab$, as was done in algorithm \ref{1st_feat_sel_alg}, and the results will be asymptotically the same; whether one uses the SVD or $\FFSVD$ on algorithm \ref{2nd_feat_sel_alg}. The latter though, which is what we present, gives a considerably faster algorithm.

\begin{algorithm}[h]
\label{2nd_feat_sel_alg}
\SetAlgoLined
\KwIn{$\Ab\in\R^{n\times d}$ for $n$ points and $d$ features, number of clusters $k$, parameter $\varepsilon\in(0,1/3)$}
\KwOut{$\Abt\in \R^{n\times r}$, with $r=O\left(k\log(k)/\varepsilon^2\right)$}
\begin{enumerate}
  \item Let $\Zb=\FFSVD(\Ab,k,\varepsilon)$; where $\Zb\in\R^{d\times r}$ \Comment{Lemma \ref{FFSVD_lem}}
  \item Let $r=c_1\cdot4k\cdot\ln(200k)/\varepsilon^2$ \Comment{$c_1$ a sufficiently large constant}
  \item Let $[\Omb,\Sb]=\RanSam(\Zb,r)$; where $\Omb\in\{0,1\}^{d\times r}$ and $\Sb\in\R^{r\times r}$ \Comment{Lemma \ref{lem_1}}
\end{enumerate}
 \Return $\Abt=\Ab\cdot(\Omb\Sb)\in\R^{n\times r}$, which consists of $r$ rescaled columns of $\Ab$
 \caption{Randomized Feature Selection, Based on $\FFSVD$ \cite{BZMD14}}
\end{algorithm}

\begin{Thm}
\label{qual_appr_thm_2}
\textit{Let $\Ab\in\R^{n\times d}$ and $k$ be the inputs of the $k$-means clustering problem, and let $\varepsilon\in(0,1/3)$. By using algorithm \ref{2nd_feat_sel_alg}, construct $\Abt\in\R^{n\times r}$ in $O(ndk/\varepsilon+k\ln(k)/\varepsilon^2\log(k\ln(k)/\varepsilon))$ time, with $r=O(k\log(k)/\varepsilon^2)$. Run any $\gamma$-approximation $k$-means algorithms with failure probability $\delta_\gamma$ on $\Abt$, $k$, and construct $\Xb_{\gamt}$. Then}
$$ \Pr\left[\|\Ab-\Xb_{\gamt} \Xb_{\gamt}^T\Ab\|_F^2\ \leq \big(1+(2+\varepsilon)\gamma\big)\|\Ab-\Xb_{opt}\Xb_{opt}^T\Ab\|_F^2\right] \geq \frac{1}{5}-\delta_\gamma. $$
\end{Thm}

$\ind$  The following discussion resembles the one in \cref{1st_feat_sel_alg}, though we give it for completeness and to introduce an alternative but equivalent characterization. Theorem \ref{qual_appr_thm_2} formally argues that the clustering obtained in a lower dimensional space will be close enough to that which would be obtained after running the $k$-means method in the original higher dimension, i.e.
\begin{equation}
\label{appr_sol_2nd_alg}
  \Ff(\Pp,\Cc_{\gamt}) \leq \big(1+(2+\varepsilon)\gamma\big)\cdot\Ff(\Pp,\Cc_{opt}) 
\end{equation}
is achieved with (relatively) high probability for $\Cc_{\gamt}$ the partition obtained after running the $\gamma$-approximation $k$-means algorithm on the lower dimensional space. Within the approximation factor $\big(1+(2+\varepsilon)\gamma\big)$, the term $\gamma\geq1$ is due to the fact that the $k$-means algorithm is performed in the lower dimensional space and does not recover the optimal partition, and the $(2+\varepsilon)$ factor is an artifact of running $k$-means in the lower dimensional space.\\
$\ind$ We give an overview of the proof. From lemma \ref{FFSVD_lem} we have that $\Eb=\Ab(\Ib_d-\Zb\Zb^T)$ and that $(\Eb\Zb)^T=\bold{0}_{k\times n}$, thus
\begin{itemize}
  \item $(P_{\Xb_{\gamt}^{\perp}}\cdot\Ab\Zb\Zb^T)(P_{\Xb_{\gamt}^{\perp}}\cdot\Eb)^T=(P_{\Xb_{\gamt}^{\perp}}\cdot\Ab\Zb)\cdot(\Zb^T\Eb^T)\cdot P_{\Xb_{\gamt}^{\perp}}^T = \bold{0}_{n\times n}$
  \item $(P_{\Xb_{\gamt}^{\perp}}\cdot\Ab\Zb\Zb^T)+(P_{\Xb_{\gamt}^{\perp}}\cdot\Eb) = P_{\Xb_{\gamt}^{\perp}}\cdot(\Ab\Zb\Zb^T+\Eb) = P_{\Xb_{\gamt}^{\perp}}\cdot\Ab = \Ab-\Xb_{\gamt}\Xb_{\gamt}^T\Ab$
\end{itemize}
so by the Pythagorean theorem \ref{mat_pyth_thm} it follows that
$$ \|\Ab-\Xb_{\gamt}\Xb_{\gamt}^T\Ab\|_F^2 = \overbrace{\|P_{\Xb_{\gamt}^{\perp}}\cdot\Ab\Zb\Zb^T\|_F^2}^{\theta_1^2}+\overbrace{\|P_{\Xb_{\gamt}^{\perp}}\cdot\Eb\|_F^2}^{\theta_2^2}. $$
$\ind$ To bound $\theta_2^2$, as was done in the term $\eta_2^2$ of algorithm \ref{2nd_feat_sel_alg} in \cref{1st_sel_alg_sec}, we may drop the $P_{\Xb_{\gamt}^{\perp}}$ term within the Frobenius norm. Then, by applying Markov's inequality\footnote{\textbf{Markov's inequality}: For a non-negative r.v. $Y$ with expectation $\E[Y]$, $\Pr\left[t\cdot\E[Y]\right]\geq1-t^{-1}$ for all $t>0$.} on \eqref{FFSVD_lem_eq}; on the nonnegative random variable $Y=\|\Eb\|_F^2-\|\Ab-\Ab_k\|_F^2$, we obtain
\begin{equation}
\label{exp_E_bound}
  \Pr\Big[\|\Eb\|_F^2\leq(1+100\varepsilon)\cdot\|\Ab-\Ab_k\|_F^2\Big]\leq0.99 \ .
\end{equation}
By the optimality of the SVD \eqref{opt_A_k}, and the fact that rank$(\Xb_{opt}\Xb_{opt}^T\Ab)\leq k$, it follows that
$$ \theta_2^2 \leq (1+100\varepsilon)\cdot\|\Ab-\Ab_k\|_F^2 \leq (1+100\varepsilon)\cdot\|\Ab-\Xb_{opt}\Xb_{opt}^T\Ab\|_F^2 = (1+100\varepsilon)\cdot F_{opt}$$
i.e. \underline{$\theta_2^2 \leq (1+100\varepsilon)\cdot F_{opt}$} with probability $0.99$.\\
$\ind$ For $\theta_1$, one needs to invoke the lemma \ref{lem_3}, the triangle inequality, drop $P_{\Xb_{\gamt}^{\perp}}$ as was done for $\theta_2^2$, use the strong submultiplicativity property, and replaces $\Xb_{opt}$ with $\Xb_{\gamt}$; while also introducing a $\sqrt{\gamma}$ multiplicative factor --- $\Xb_{\gamt}$ gives a $\gamma$-approximation to the optimal $k$-means clustering $\Abt=\Ab\cdot(\Omb\Sb)=\Ab\Pib$; thus any other $n\times k$ indicator matrix satisfies
\begin{equation}
\label{ident_prop_ind_matr}  
  \|(\Ib_n-\Xb_{\gamt}\Xb_{\gamt}^T)\Abt\|_F^2 \leq \gamma\cdot\min_{\Xb\in\X}\Big\{\|(\Ib_n-\Xb\Xb^T)\Abt\|_F^2\Big\} \leq \gamma\cdot\|(\Ib_n-\Xb_{opt}\Xb_{opt}^T)\Abt\|_F^2.
\end{equation}
All in all, by the above and assuming that $\gamma\leq1$ (in order to compute the failure probability) we obtain
$$ \theta_1\leq \sqrt{\gamma}\|(\Ib_m-\Xb_{opt}\Xb_{opt}^T)\Abt\|_F\cdot\|(\Zb\Pib)^{\dagger}\|_2+\|\tilde{\Eb}\|_F $$
and by using lemmas \ref{lem_1} and \ref{lem_2} with appropriate parameters $\delta$ and $\varepsilon$, after performing ``elementary'' algebra, we conclude that
$$ \underline{\theta_1^2 \leq \big(\sqrt{2}+94\varepsilon\big)^2\cdot\gamma F_{opt} \leq (2+3900\varepsilon)\cdot\gamma F_{opt}}. $$
Putting the two bounds together and under the assumption that $\gamma\leq1$, we get
$$ \|\Ab-\Xb_{\gamt}\Xb_{\gamt}^T\Ab\|_F^2 = \theta_1^2+\theta_2^2 \leq \big((2+3900\varepsilon)\cdot\gamma+(1+100\varepsilon)\big)\cdot F_{opt} \leq \big(1+(2+4000\varepsilon)\cdot\gamma\big)\cdot F_{opt}. $$
Again by selecting the appropriate parameters, and applying the union bound on lemma \ref{lem_2} and Markov's inequality on \eqref{FFSVD_lem_eq}, we get a failure probability of $0.8+\delta_{\gamma}$. Hence, we attain the bound in theorem \ref{qual_appr_thm_2} with probability at least $0.2-\delta_{\gamma}$.

\section{Randomized Feature Extraction Algorithms}
\label{feat_extr_algs_sec}

\subsection{First Randomized Feature Extraction Algorithm --- Random Projection}
\label{1st_extr_alg_sec}

$\ind$ Our first feature extraction algorithm, relies on projecting the matrix $\Ab\in\R^{n\times d}$ which is comprised of the points $\Pp$, into a space of dimension $r=O(k/\varepsilon^2)<d$, in time $O(nd\ceil{\varepsilon^{-2}k/\log(n)})$, such that the objective value of the optimal $k$-partition $\Cc_{opt}$ of $\Pp$ is preserved within a factor of $2+\varepsilon$, with constant probability.\\
$\ind$ The projection takes place by post-multiplying $\Ab$ by a matrix $\Pib\in\{\rpm1/\sqrt{r}\}^{d\times r}$ whose entries are essentially realizations of a Rademacher random variable, which are then rescaled by $1/\sqrt{r}$. That is, $\Pib_{ij}=1/\sqrt{r}$ with probability $0.5$; and $\Pib_{ij}=-1/\sqrt{r}$ with probability $0.5$. It takes $O(nk/\varepsilon^2)$ time to generate $\Pib$, and the multiplication is done by the \textit{mailman algorithm} \cite{LZ09}, which will take $O(nd\ceil{\varepsilon^{-2}k/\log(n)})$ time. For a short explanation of the mailman algorithm, refer to appendix \ref{mailman_app}. Algorithm \ref{1st_feat_extr_alg} also resembles ideas from \cite{Ach01}, \cite{Ach03}, which seems to be one of the first algorithms to propose a projection matrix of this kind, while simultaneously considering how the projection; i.e. matrix multiplication, could be done relatively fast. The projection matrix $\tilde{\Pib}$ from \cite{Ach01} considers entries
$$ \tilde{\Pib}_{ij} = \begin{cases} \sqrt{3} \qquad \ \ \ \text{ w.p. } 1/6 \\ -\sqrt{3} \qquad \ \text{ w.p. } 1/6 \\ 0 \qquad \ \ \ \ \ \ \text{ w.p. } 2/3 \end{cases} $$
so that it can construct $\Ab\tilde{\Pib}$ fast, where $\tilde{\Pib}$ is in fact a \textit{Johnson-Lindenstrauss transform}.

\begin{algorithm}[h]
\label{1st_feat_extr_alg}
\SetAlgoLined
\KwIn{$\Ab\in\R^{n\times d}$ for $n$ points and $d$ features, number of clusters $k$, parameter $\varepsilon\in(0,1/3)$}
\KwOut{$\Abt\in \R^{n\times r}$, with $r=O\left(k/\varepsilon^2\right)$ artificial features}
\begin{enumerate}
  \item Set $r\gets c_2\cdot k/\varepsilon^2$\Comment{$c_2$ a sufficiently large constant}
  \item Generate a random matrix $\Pib\in\{\rpm1/\sqrt{r}\}^{d\times r}$, s.t. $\forall(ij)\in\N_d\times\N_r$ (i.i.d.)
  $$ \Pib_{ij} = \begin{cases} +1/\sqrt{r} \qquad \text{w.p. } 1/2 \\ -1/\sqrt{r} \qquad \text{w.p. } 1/2 \end{cases} $$
  \item Let $\Zb=\FFSVD(\Ab,k,\varepsilon)$; where $\Zb\in\R^{d\times r}$
  \item Compute $\Abt=\Ab\Pib\in\R^{n\times r}$, using the \textit{Mailman Algorithm} \cite{LZ09}
\end{enumerate}
 \Return $\Abt$, which consists of $r$ extracted features of each row of $\Ab$
 \caption{{\small Randomized Feature Extraction, Based on Random Projections} \cite{BZD10}, \cite{BZMD14}}
\end{algorithm}

\begin{Thm}
\label{qual_appr_thm_3}
\textit{Let $\Ab\in\R^{n\times d}$ and $k$ be the inputs of the $k$-means clustering problem, and let $\varepsilon\in(0,1/3)$. By using algorithm \ref{1st_feat_extr_alg}, construct $\Abt\in\R^{n\times r}$ with $r=O(k/\varepsilon^2)$ in $O(nd\ceil{\varepsilon^{-2}k/\log(d)})$ time. Run any $\gamma$-approximation $k$-means algorithms with failure probability $\delta_\gamma$ on $\Abt$, $k$, and construct $\Xb_{\gamt}$. Then}
$$ \Pr\left[\|\Ab-\Xb_{\gamt} \Xb_{\gamt}^T\Ab\|_F^2\ \leq \big(1+(1+\varepsilon)\gamma\big)\|\Ab-\Xb_{opt}\Xb_{opt}^T\Ab\|_F^2\right] \geq 0.96-\delta_\gamma. $$
\end{Thm}

$\ind$ Loosely speaking, in algorithm \ref{1st_feat_extr_alg} it suffices to create roughly $O(k)$ new features via a random projection, and then run some (approximate) $k$-means algorithm on the resulting $\Abt$, in order to obtain
\begin{equation}
\label{qual_of_appr_feat_extr_algs}
  \Ff(\Pp,\Cc_{\gamt}) \leq \big(1+(1+\varepsilon)\gamma\big)\cdot\Ff(\Pp,\Cc_{opt})
\end{equation}
where the reasoning of the approximation factor $\big(1+(1+\varepsilon)\gamma\big)$ is analogous to that of \eqref{appr_sol_2nd_alg}.\\
$\ind$ Let us go over the main steps of the proof. Similarly to the proofs of theorems \ref{qual_appr_thm_1} and \ref{qual_appr_thm_2}, the objective is separably additive into orthogonal pairs
$$ \|\Ab-\Xb_{\gamt}\Xb_{\gamt}^T\Ab\|_F^2 = \overbrace{\|P_{\Xb_{\gamt}^{\perp}}\cdot\Ab_k\|_F^2}^{\xi_1^2}+\overbrace{\|P_{\Xb_{\gamt}^{\perp}}\cdot\Ab_{\rho-k}\|_F^2}^{\xi_2^2}. $$
Since the terms $\xi_2^2$ and $\eta_2^2$ from \cref{1st_sel_alg_sec} are the same, we conclude that \underline{$\xi_2^2\leq F_{opt}$}. To bound $\xi_1^2$, we use the same tools that were used for bounding $\theta_2^2$, with the exception of now using lemmas \ref{lem_5} with $\Yb=(\Ib_n-\Xb_{opt}\Xb_{opt}^T)\Ab$ and \ref{lem_6} part 2; instead of lemma \ref{lem_3}, to get identity \eqref{ident_prop_ind_matr} and
$$ \xi_1\leq\sqrt{\gamma}(1+5.5\varepsilon)\cdot\|(\Ib_n-\Xb_{opt}\Xb_{opt}^T)\Ab\|_F \quad \implies \quad \underline{\xi_1^2\leq\sqrt{\gamma}(1+15\varepsilon)\cdot\|(\Ib_n-\Xb_{opt}\Xb_{opt}^T)\Ab\|_F^2}. $$
Furthermore, the parameters are scaled accordingly and the hyper-parameter $c_2$ is set to $c_2=3330\cdot15^2$, in order to get a failure probability of $0.04+\delta_\gamma$ after applying the union bound. The probability of attaining the desired approximation accuracy, is therefore $0.96-\delta_\gamma$.\\

$\ind$ Compared to the brief discussion in \cref{k_means_subsec} where we described how the Johnson-Lindenstrauss could be applied, theorem \ref{qual_appr_thm_3} argues that through algorithm \ref{1st_feat_extr_alg} a much smaller dimension suffices in order to preserve the optimal clustering in the data. We also point out that the pairwise distances after applying algorithm \ref{1st_feat_extr_alg} are not proven to be preserved, where instead the analysis shows that if the spectral information of certain matrices is preserved; then the $k$-means clustering is also preserved.

\subsection{Second Randomized Feature Extraction Algorithm --- Approximate SVD}
\label{2nd_extr_alg_sec}

$\ind$ The final algorithm we present is based on the ``Approximate SVD'', which method and proof techniques are similar to those from \cite{DFKV99}, where in the proposed $k$-means algorithm the faster $\FFSVD$ algorithm is used instead of the exact deterministic SVD. The quality-of-approximation result is given in theorem \ref{qual_appr_thm_4}, in which the dimension of the data points is reduced from $d$ to the number of clusters $r=k$.

\begin{algorithm}[h]
\label{2nd_feat_extr_alg}
\SetAlgoLined
\KwIn{$\Ab\in\R^{n\times d}$ for $n$ points and $d$ features, number of clusters $k$, parameter $\varepsilon\in(0,1)$}
\KwOut{$\Abt\in \R^{n\times r}$, with $r=k$ artificial features}
\begin{enumerate}
  \item Let $\Zb=\FFSVD(\Ab,k,\varepsilon)$; where $\Zb\in\R^{d\times k}$ \Comment{Lemma \ref{FFSVD_lem}}
  \item Compute $\Abt=\Ab\Zb\in\R^{n\times k}$
\end{enumerate}
 \Return $\Abt$, which consists of $r=k$ extracted features of each row of $\Ab$
 \caption{Randomized Feature Extraction, Based on Approximate SVD \cite{BZMD14}}
\end{algorithm}

\begin{Thm}
\label{qual_appr_thm_4}
\textit{Let $\Ab\in\R^{n\times d}$ and $k$ be the inputs of the $k$-means clustering problem, and let $\varepsilon\in(0,1)$. By using algorithm \ref{2nd_feat_extr_alg}, construct $\Abt\in\R^{n\times k}$ in $O(ndk/\varepsilon)$ time. Run any $\gamma$-approximation $k$-means algorithms with failure probability $\delta_\gamma$ on $\Abt$, $k$, and construct $\Xb_{\gamt}$. Then}
$$ \Pr\left[\|\Ab-\Xb_{\gamt} \Xb_{\gamt}^T\Ab\|_F^2\ \leq \big(1+(1+\varepsilon)\gamma\big)\|\Ab-\Xb_{opt}\Xb_{opt}^T\Ab\|_F^2\right] \geq 0.99-\delta_\gamma. $$
\end{Thm}

$\ind$ Note that the approximation result of theorem \ref{qual_appr_thm_4} is the same as that of theorem \ref{qual_appr_thm_3}; i.e. \eqref{qual_of_appr_feat_extr_algs}, with slightly higher probability. Again, the reasoning of the approximation factor $\big(1+(1+\varepsilon)\gamma\big)$ is analogous to the one given for \eqref{appr_sol_2nd_alg}.\\
$\ind$ As with all other algorithms we discussed, the first step is to separate the objective
$$ \|\Ab-\Xb_{\gamt}\Xb_{\gamt}^T\Ab\|_F^2 = \overbrace{\|P_{\Xb_{\gamt}^{\perp}}\cdot\Ab\Zb\Zb^T\|_F^2}^{\pi_1^2}+\overbrace{\|P_{\Xb_{\gamt}^{\perp}}\cdot\Eb\|_F^2}^{\pi_2^2}. $$
which is the same expression we obtained in the analysis of algorithm \ref{2nd_feat_sel_alg}, and then bound each of the two terms. After all, both algorithms \ref{2nd_feat_sel_alg} and \ref{2nd_feat_extr_alg} are based on $\FFSVD$. It follows through that \eqref{exp_E_bound} is obtained, hence with probability $0.99$ we have \underline{$\pi_2^2 \leq (1+100\varepsilon)\cdot F_{opt}$}.\\
$\ind$ For the first term we have
$$ \pi_1 \leq \|(\Ib_n-\Xb_{\gamt}\Xb_{\gamt}^T)\Ab\Zb\|_F \leq \sqrt{\gamma}\|(\Ib_n-\Xb_{opt}\Xb_{opt}^T)\Ab\Zb\|_F \leq \sqrt{\gamma}\|(\Ib_n-\Xb_{opt}\Xb_{opt}^T)\Ab\|_F $$
where in the first inequality we apply the submultiplicativity property and use the fact that $\|\Zb^T\|_2=1$, the second is a similar argument to what we saw in the proof of theorem \ref{qual_appr_thm_2}, and the third inequality again follows from the submultiplicativity property and the fact that $\|\Zb\|_2=1$. Therefore, we have
$$ \underline{\pi_1^2 \leq \big(\sqrt{\gamma}\|(\Ib_n-\Xb_{opt}\Xb_{opt}^T)\Ab\|_F\big)^2 = \gamma F_{opt}} $$
and the overall error for $\gamma\geq1$ is
$$ \|\Ab-\Xb_{\gamt}\Xb_{\gamt}^T\Ab\|_F^2 \leq \pi_1^2+\pi_2^2 \leq \gamma F_{opt} + (1+100\varepsilon)\cdot F_{opt} \leq \big(1+(1+100\varepsilon)\big)\cdot\gamma F_{opt}. $$
By definition \ref{gam_appr_def} and using the union bound on lemma \ref{FFSVD_lem}, the failure probability is $0.01+\delta_\gamma$, thus the probability of success is $0.99-\delta_\gamma$. By rescaling $\varepsilon$ accordingly; we get the approximation factor of $\big(1+(1+\varepsilon)\gamma\big)$.

\section{Summary and Concluding Remarks}
\label{sum_concl_sec}

$\ind$ In this manuscript we saw four of the first provably accurate approximations for the $k$-means clustering problem. We did not discuss any implementations of these, though evaluation experiments may be found in the references from which the algorithms were taken, e.g. \cite{BZMD14}. These are also being compared with other dimensionality reduction techniques, such as the Laplacian scores \cite{HCN05}. A summary of the results we discussed is provided in the table below.

\begin{center}
\label{table_results}
\begin{tabular}{ |p{.7cm}||p{1.8cm}|p{2.2cm}|p{2.2cm}|p{2.7cm}|p{2.55cm}|p{1.5cm}| }
\hline
\multicolumn{7}{|c|}{\textbf{Summary of the results we presented}} \\
\hline
 \textbf{Alg.} & \textbf{Reference} & \textbf{Description} & \textbf{Dimensions} & \textbf{Time $O(\cdot)$} & \textbf{Appr. Factor} & \textbf{Probab.}\\
\hline
 \ref{1st_feat_sel_alg} & \cite{BDM09} & Lev. Scores & {\small$\Theta(k\log(k\varepsilon)/\varepsilon^2)$} & $nd\cdot\min\{n,d\}$ & $\big(1+(2+\varepsilon)\gamma\big)$ & $0.5-\delta_\gamma$ \\
\hline
 \ref{2nd_feat_sel_alg} & \cite{BDM14} & Ran. Sampl. & {\small$O(k\log(k)/\varepsilon^2)$} & Theorem \ref{qual_appr_thm_2} & $\big(1+(2+\varepsilon)\gamma\big)$ & $0.5-\delta_\gamma$ \\ 
\hline
 \ref{1st_feat_extr_alg} & \cite{BDM14} & Ran. Proj. & {\small$O(k/\varepsilon^2)$} & $nd\ceil{\varepsilon^{-2}k/\log(n)}$ & $\big(1+(1+\varepsilon)\gamma\big)$ & $0.96-\delta_\gamma$ \\
\hline
 \ref{2nd_feat_extr_alg} & \cite{BDM14} & Appr. SVD & $k$ & $ndk/\varepsilon$ & $\big(1+(1+\varepsilon)\gamma\big)$ & $0.99-\delta_\gamma$ \\
\hline
\end{tabular}
\end{center}
\vspace{5mm}

$\ind$ Since the first algorithm in the series of papers we summarized is based on leverage scores, it is worth pointing their importance in randomized matrix algorithms and randomized numerical linear algebra. They have been the key structural quantity which helped understand and bridge the ``theory-practice gap'' between theoretical work on randomized algorithms for large matrices, and applications, both numerical implementation and data-analysis applications \cite{DM10}. To mention a few applications, they are used for low-rank matrix approximations, solving over-constrained least squares and the column subset selection problem, and speeding up Laplacian solvers \cite{SS11}. For more details on these applications, refer to \cite{DM10} and the references therein.\\
$\ind$ In the case where one would want to accelerate algorithm \ref{1st_feat_sel_alg}, he or she could use approximate leverage scores instead of computing the top $k$ right singular vectors of $\Ab$. This would of course result in a looser approximation of the $k$-means problem, though in practice speed may be a bigger issue. The leverage scores can be approximated in time $O(nd\log(n)/\varepsilon^2))$ for an error parameter $\varepsilon$ \cite{DMMW12}. It would be interesting to see how good the leverage scores approximations are if we simply take $\Zb=\FFSVD(\Ab,k,\varepsilon)$ and define the approximate scores as $\hat{\ell}_i=\|(\Zb_{\rho})_{(i)}\|_2^2$, as well as how good an approximation we would get to the $k$-means problems if these were used for importance sampling.\\

$\ind$ A few papers which we did not discuss are \cite{BMD08} which solves the principal component problem in a similar fashion to our first two algorithms, and \cite{WLRT08} which suggest that a sub-sampled randomized Fourier transform can be used for the design of a provably accurate feature extraction algorithm for $k$-means clustering. There are dozens of papers which one could relate to what we discussed, and it is not possible to list them all.\\

$\ind$ Ever since the results we reviewed were first published, there have been substantial improvements \cite{CEMMP15}. One of the main ideas used in \cite{CEMMP15} is to use a reduction for results over $\|\cdot\|_F$, to apply them to $\|\cdot\|_2$. Column selection and random projection constructions are presented for dimensionality reduction to construct $\Abt=\Ab\Pib\in\R^{n\times r}$ where $\Pib\in\R^{d\times r}$ is respectively a diagonal matrix that selects and re-weights the columns of $\Ab$, or a random Johnson-Lindenstrauss matrix. These $\Abt$ matrices are \textbf{rank-$k$ projection-cost preserving sketches with two-sided error} with error $\varepsilon$. That is, for error $\varepsilon\in[0,1)$; for all rank-$k$ orthogonal projection matrices $\Pib\in\R^{n\times n}$
$$ (1-\varepsilon)\|\Ab-\Pib\Ab\|_{\xi}^2\leq\|\Abt-\Pib\Abt\|_{\xi}^2+c\leq(1+\varepsilon)\|\Ab-\Pib\Ab\|_{\xi}^2 $$
for some $c\geq0$ that may depend on $\Ab$ and $\Abt$, but not on $\Pib$. The reduction is going from $\xi=F$ to $\xi=2$. In addition to many of the ideas we have already seen, a key to their constructions is to use orthogonal projections, which by the submultiplicativity property implies that the Frobenius norm can only decrease, since all singular values of such projections are either $0$ or $1$.\\
$\ind$ More recently results may be found in \cite{MMR19} and \cite{BBCAGS19}.


\bibliographystyle{alpha}
\bibliography{refs.bib}

\appendix

\section{The Mailman Algorithm}
\label{mailman_app}

$\ind$ We give a brief explanation, as well as the intuition and reasoning of the name behind the \textit{mailman algorithm}, which is used in algorithm \ref{1st_feat_extr_alg} to speed up the matrix multiplication.\\
$\ind$ Speeding up matrix computations is a rich topic in many areas of mathematics, engineering and computer science (as we have already seen throughout this manuscript), and at their core, the problems which have probably been studied the most; are matrix-matrix and matrix-vector multiplication. The mailman algorithm deals with matrix-vector multiplication which can then be extended to matrix-matrix multiplication, where the matrix acting on the vector takes entries from a \textit{finite alphabet} $\Sigma$. Prior to this work, the case of dealing with matrices over finite fields had been extensively studied; though here the vector can take real values.\\
$\ind$ Intuitively, the algorithm multiplies $A\in\Sigma^{m\times n}$ and $x\in\R^{n}$ in a manner that resembles the following ``principle'' when distributing mail `\textit{first sorting the letters by address and then delivering them}'. Viewing this multiplication as $Ax=\sum_{i=1}^nA^{(i)}x_i$, the columns $A^{(i)}$ correspond to the \textit{addresses} and the entries $x_i$ to a \textit{letter addressed to it}, where the sum is equivalent to \textit{the effort of walking to $A^{(i)}$ and delivering $x_i$}. Naively, this addition corresponds to delivering each letter one by one --- results in the naive matrix-vector multiplication which takes $O(mn)$ time --- though this is not what an efficient mailman would do (specially during a pandemic such as coronavirus)! The mailman's strategy would be to:
\begin{enumerate}
  \item first arrange his or her letters according to the shortest route including \textit{all} houses
  \item walk the route visiting each house only once; regardless of how many letters should be delivered to it (possibly none).
\end{enumerate}
$\ind$ The mailman algorithm applies this strategy, by decomposing $A$ and exploiting the fact that matrix-matrix multiplication is associative. The decomposition we consider is mainly for performing matrix-matrix multiplication rather than the ``simpler'' matrix-vector multiplication. For simplicity we consider $\Sigma=\{0,1\}$ and assume that $m=\log_2(n)$. There are $2^m=n$ distinct possible columns which each column of $A$ could be equal to, which we append in a matrix $U_n\in\Sigma^{m\times n}$. The matrix $U_n$ is referred to as the \textit{universal columns matrix}. It follows that for every $j\in\N_n$ there exists a \textit{unique} $i\in\N_n$ for which $A^{(j)}=U_n^{(i)}$. We then define the \textit{correspondence matrix} (in the sense that it indicates which letter would be delivered to which address) $P_{ij}=\mathbbm{1}\{U_n^{(i)}=A^{(j)}\}$\footnote{$\mathbbm{1}\{\alpha=\beta\}=1$ if $\alpha=\beta$, and $\mathbbm{1}\{\alpha=\beta\}=0$ if $\alpha\neq\beta$.}, which $P\in\{0,1\}^{n\times n}$ has exactly $n$ non-zero entries; one in each row and each column. Row $P_{(j)}$ corresponds to the column $A^{(j)}$, and column $P^{(i)}$ to the column $U_n^{(i)}$. We then have the decomposition $A=UP$, where $P$ encodes all the information of $A$ since $U_n$ is fixed.\\
$\ind$ Now that we have decomposed $A$, we first apply $x$ to $P$, which is analogous to 1. arranging the letters The second step of applying $U$ to $(Px)$ is analogous to 2. walking the route. Here is how we take advantage of the associativity of matrix-matrix multiplication
\begin{align*}
  (U_nP)_{ij} &= \sum\limits_{l=1}^n(U_n)_{il}\cdot P_{lj}\\
  &= \sum\limits_{l=1}^n(U_n^{(l)})_i\cdot \mathbbm{1}\{U_n^{(l)}=A^{(j)}\}\\
  &= (A^{(j)})_i = A_{ij}.
\end{align*}
Since nnzr$(P)=n$, it takes $O(n)$ operations to compute $Px$, and by the construction of $U_n$; applying it to any vector requires $O(n)$ operations. All in all, it takes $O(n+n)=O(n)$ operations to compute $Ax=U_n(Px)$. For the general case and $m>\log_{|\Sigma|}(n)$, please refer to \cite{LZ09}.

\subsection{Applying the Algorithm}

$\ind$ As mentioned above in the case of $\Sigma=\{0,1\}$, the universal column matrix is made up of all rows of length $m=\log_2(n)$ over $\Sigma$, which may be constructed by assigning to $U_n^{(i)}$ the binary representation of $i\in\N_n-1$. Though not mentioned in \cite{LZ09}, this is closely resembles a construction of the generator matrix of certain Hamming and Reed-Muller error-correcting codes! More precisely, if we remove the all zeros column vector $U_n^{(1)}$, we get the transpose of the generator matrix of a Hamming code. In the binary case, if the all-zeros vector $U_n^{(1)}$ is replaced with the all-ones vector, we get the transpose of the generator matrix of certain first order Reed-Muller codes. The transpose is for the common convention where codewords are represented by row rather than column vectors.\\
$\ind$ The matrix $U_n\in\Sigma^{\log_2(n)\times n}$ for $\Sigma=\{0,1\}$ can also be constructed recursively as follows
$$ U_1=\begin{pmatrix} 1 & 0 \end{pmatrix} \qquad \text{ and } \qquad U_n = \begin{pmatrix} \bold{0}_{n/2}^T & | & \bold{1}_{n/2}^T \\ - \ -  & - & - \ -  \\ U_{n/2}^T & | & U_{n/2}^T \end{pmatrix} \in \Sigma^{\log_2(n)\times n}. $$
Applying $U_n$ to any vector $z^T=\begin{pmatrix}z_1^T & z_2^T\end{pmatrix}$ of length $n$ with $z_1,z_2$ each of length $n/2$, takes less than $4n$ operations. The product $U_n\cdot z$ can be computed recursively
$$ U_n\cdot z = \begin{pmatrix} \bold{0}_{n/2}^T & | & \bold{1}_{n/2}^T \\ - \ - & - & - \ - \\ U_{n/2}^T & | & U_{n/2}^T \end{pmatrix} \cdot \begin{pmatrix} z_1 \\ - \\ z_2 \end{pmatrix} = \begin{pmatrix} \bold{0}_{n/2}^T\cdot z_1 + \bold{1}_{n/2}^T\cdot z_2 \\ - - - - - - - - - \\ U_{n/2}\cdot(z_1+\ z_2) \end{pmatrix} = \begin{pmatrix} \bold{1}_{n/2}^T\cdot z_2 \\ - - - - - \\ U_{n/2}\cdot z \end{pmatrix} $$
where computing $\bold{0}_{n/2}^T\cdot z_1 + \bold{1}_{n/2}^T\cdot z_2$ takes no more than $2n$ operations. For $T(n)$ the operation count required for $U_n\cdot z$, it follows that
$$ T(2)=2 \ \ \text{ and } \ \ T(n)=T(n/2)+2n \quad \implies \quad T(n)\leq 4n $$
where $\bold{0}_{n/2}^T\cdot z_1$ could of course be neglected.\\
$\ind$ The construction of $P$ is done by reading $A$, and setting $P_{ij}=1$ only when $A^{(j)}$ gives the binary representation of $i-1$. This is done in $O(mn)$ steps. Constructing $z=Px$ then takes an additional $O(n)$ operations. Since $m<n$, we need $O(mn+n+4n)=O(n)$ operations to compute $Ax$.\\

$\ind$ In the case of $A\in\Sigma^{m\times n}$ for $S\coloneqq |\Sigma|$ such that $2<S<\infty$, the matrix can be decomposed similarly and applied to any vector $x\in\R^{n}$ in $O(mn\log(S)/\log(n))$ time. Again, just as in the case of Hamming codes over higher alphabets (though here we consider any finite $\Sigma$, not just finite fields), the universal column matrix $U_n$ encodes all possible strings over $\Sigma=\{\sigma_1,\cdots,\sigma_S\}$
$$ U_1=\begin{pmatrix} 1 & 0 \end{pmatrix} \qquad \text{ and } \qquad U_n = \begin{pmatrix} \sigma_1\cdot\bold{1}_{n/S}^T & | & \cdots & | & \sigma_S\cdot\bold{1}_{n/S}^T \\ - \ - \ - & - & - \ - & - & - \ - \ -  \\ U_{n/S}^T & | & \cdots & | & U_{n/S}^T \end{pmatrix} \in \Sigma^{\log_S(n)\times n}\ . $$
The product $U_n\cdot z$ itself takes $O(n)$ time. For $P$, we set $P_{ij}=1$ if and only if $A^{(j)}$ represents $i-1$ in base $S$, under the transformation $\tau:\Sigma\to(\N_S-1)$ such that $\tau:\sigma_l\mapsto(l-1)$.\\

$\ind$ Furthermore, the mailman algorithm has applications to other dimensionality reduction algorithms \cite{LZ09}. For instance, it can be combined with earlier results by Achlioptas \cite{Ach01}, \cite{Ach03} we encountered in \cref{1st_extr_alg_sec}, to speed up the corresponding embedding even more.

\end{document}